\documentclass[runningheads]{llncs}
\usepackage{graphicx}
\usepackage{subfigure} 
\usepackage{multirow}
\usepackage{array}
\usepackage{amssymb}
\usepackage{color}
\usepackage{algorithm}
\usepackage{algorithmic}
\usepackage{bm}
\begin{document}
\title{ A Diversity-Enhanced and Constraints-Relaxed Augmentation for Low-Resource Classification}
%
%
\author{Guang Liu \and Hailong Huang \and Yuzhao Mao \and Weiguo Gao \and Xuan Li \and Jianping Shen}
%
%
\institute{PingAn Life Insurance of China}
%
\maketitle              
\begin{abstract}
Data augmentation (DA) aims to generate constrained and diversified data to improve classifiers in Low-Resource Classification (LRC). Previous studies mostly use a fine-tuned Language Model (LM) to strengthen the constraints but ignore the fact that the potential of diversity could improve the effectiveness of generated data. In LRC, strong constraints but weak diversity in DA result in the poor generalization ability of classifiers. To address this dilemma, we propose a \textbf{D}iversity-\textbf{E}nhanced and \textbf{C}onstraints-\textbf{R}elaxed \textbf{A}ugmentation (\textbf{DECRA}). Our DECRA has two essential components on top of a transformer-based backbone model. 1) A \textbf{\textit{$\mathbf{k}$-$\bm \beta$ augmentation}}, an essential component of DECRA, is proposed to enhance the diversity in generating constrained data. It expands the changing scope and improves the degree of complexity of the generated data. 2) A masked language model loss, instead of fine-tuning, is used as a \textbf{regularization}. It relaxes constraints so that the classifier can be trained with more scattered generated data. The combination of these two components generates data that can reach or approach category boundaries and hence help the classifier generalize better. We evaluate our DECRA on three public benchmark datasets under low-resource settings. Extensive experiments demonstrate that our DECRA outperforms state-of-the-art approaches by 3.8\% in the overall score.

\end{abstract}
\section{Introduction}
\begin{figure*}[htbp]
\centering
\subfigure[Previous works]{
\includegraphics[width=0.45\columnwidth]{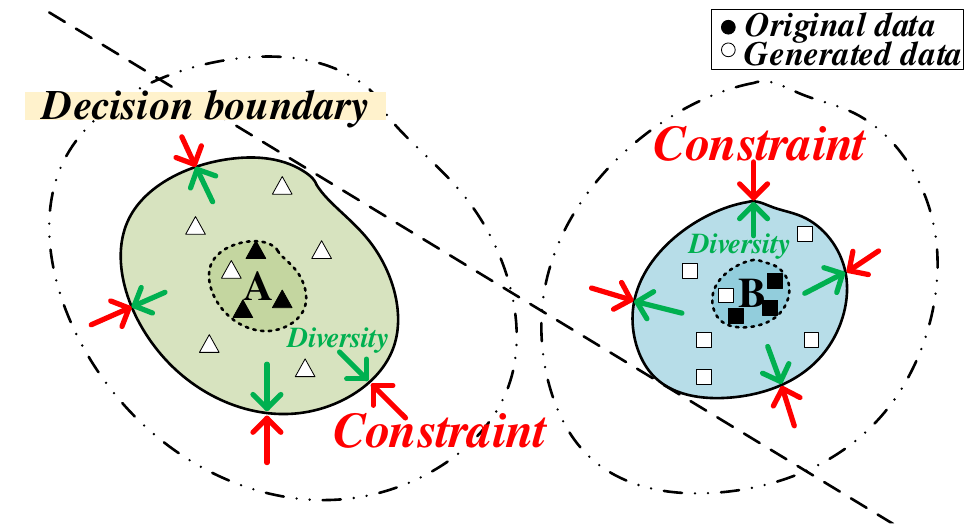}
\label{fig1:subfig1}
}
\subfigure[Ours]{
\includegraphics[width=0.45\columnwidth]{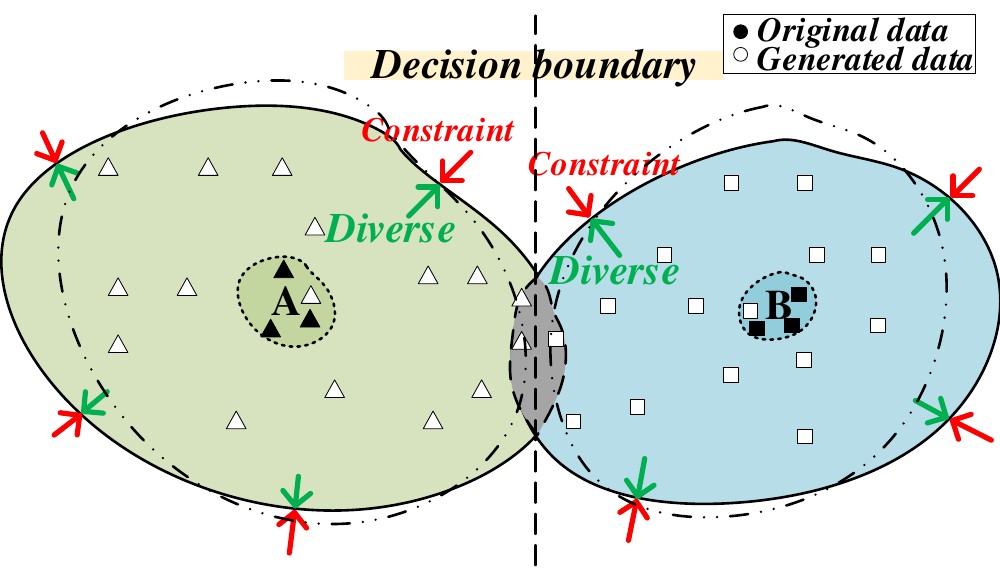}
\label{fig1:subfig2}
}
\caption{The demonstration of how augmentation works in LRC\label{fig1}. \textit{Fig~\ref{fig1:subfig1} is the augmentation with strong constraints and weak diversity. The generated data is near to the original ones. So, the classifier learns a decision boundary that generalizes poorly. Fig~\ref{fig1:subfig2} is augmentation with enhanced diversity and relaxed constraints. Ideally, the generated data will will be close to the boundary of the categories. Therefore, the classifier learns a decision boundary that has a better generalization ability.}}
\end{figure*}
Data Augmentation (DA) approaches~\cite{cubuk2019autoaugment,antoniou2017data,hernandez2018data,hu2019learning,wang2020generalizing} are often used to alleviate the thirst for labeled data in Low-Resource Classification (LRC~\cite{shleifer2019low}). Classification is essential in building intelligent systems, such as empathetic dialogue systems~\cite{young2020dialogue,bertero2016real} and medical diagnosis systems~\cite{abu2017medical,li2017multi}. In most cases, external resources, such as the similar task data~\cite{wang2020generalizing} and unlabeled data~\cite{tu2019review}, are hard to obtain or even unavailable. In low-resource settings, the greatest challenge is the expensive cost as well as the shortage of experienced experts who serve to collect and label large-scale data. Without sufficient labeled data, the deep networks tend to generalize poorly, leading to unsatisfactory performance. 

Data Augmentation (DA) in text data aims to generate constrained and diversified data to improve classifier performance~\cite{xia2019generalized,gupta2019data,howard2018universal,hu2019learning}. Ideally, we assume data generated by DA is able to present the data distribution in every category. Thus, the generated data is supposed to extend the range of labeled data which would help the classifier make better decisions~\cite{wei2019eda}. As shown in Fig.~\ref{fig1:subfig1}, constraints and diversity are two main concepts in DA. The constraints mainly pull the generated data towards the original one. The diversity in generating comes from partially changed labeled data. It pushes the generated data away from the original one. Previous researches focus on strengthening the constraints~\cite{kobayashi2018contextual,wu2019conditional,hu2019learning}, especially contextually. The fine-tuned~\cite{anaby2020not,gururangan2020don} Language Models (LM) are often used to generate contextual constraints, e.g., Bidirectional Encoder Representations from Transformers (BERT\cite{devlin2019bert}). In the meantime, additional constraints from the labels are introduced through fine-tuning, such as the Conditional BERT (CBERT~\cite{kobayashi2018contextual}), which is fine-tuned with the additional constraints on labels. CBERT overfits in the low data conditions and consequently lacks diversity in generating. To improve generalization ability, the Learning Data Manipulate for Augmentation and Weighting (LDMAW~\cite{hu2019learning}) unifies the learning targets of both the augmentation and the classification through a reinforcement learning framework.  Noticeably, LDMAW uses a BERT (fine-tuned LM) to generate data for another BERT (classifier).

As depicted in Fig.~\ref{fig1:subfig1}, previous studies often suffer from poor generalization ability in low-resource conditions~\cite{tu2019review} due to strong constraints but weak diversity in augmentation. As the Language Model (LM) tends to overfit on limited data in low-resource conditions, strong constraints are formed after fine-tuning~\cite{wu2019conditional,hu2019learning}. As a result, the generated data is pulled towards the original data. At the same time, the weakness of diversity in augmentation is often ignored. Current DA approaches mostly use the method that is identical to the masked Language Model learning in BERT~\cite{devlin2019bert}. In this method, diversity is influenced by the changing scope and degree of complexity in the generated data. The changing scope is proportional to the times of the DA applied.  In each time of augmentation, one set of maskers is generated~\cite{hu2019learning,kobayashi2018contextual}. And the masked positions are the ones to be augmented. This process results in a fixed and narrow changing scope. On the other hand, the degree of complexity is related to the amount of information used in the augmenting data in masked positions. For each masked position, routinely, one sampled tokens are applied~\cite{wu2019conditional,kobayashi2018contextual}. Therefore, it results in the low complexity of the generated data. Consequently, strong constraints but weak diversity causes the poor generalization ability in LRC.

To address the described problem, we propose a \textbf{D}iversity-\textbf{E}nhanced and \textbf{C}onstraints-\textbf{R}elaxed \textbf{A}ugmentation (\textbf{DECRA}), as displayed in Fig.~\ref{fig1:subfig2}. DECRA allows the generated data to be more scattered within the extended boundary. Our DECRA is based on the modified LDMAW~\cite{hu2019learning}, which is the state-of-the-art model in LRC. The backbone model, a simplified LDMAW, shares parameters in BERT to reduce overfitting for better generalization ability. The backbone model consists of a transformer-based encoder (TBE), a language model layer (LML) and a classification layer (CL). DECRA has two essential components based on the backbone model: $k$-$\beta$ augmentation and regularization (masked LM loss). 1) $k$-$\beta$ augmentation, an essential component in DECRA, will enhance the diversity in generating. It expands the changing scope by applying augmentation $\beta$ times and enhances the degree of complexity by using top-k tokens to augment the masked position. 2) The regularization, masked LM loss on original data, generates more relaxed constraints compared to fine-tuning. DECRA will be trained by the combination of masked LM loss and the classification loss. Our model can learn the constraints dynamically and progressively during the training process. It will processe more scattered generated data, which will reach or approach the boundary of categories, to achieve better generalization ability. Therefore, enhanced diversity as well as relaxed constraints help to generate data more scattered within the extended boundary. Trained with the labeled and generated data, the classifier will make better decisions and consequently achieve better generalization ability in LRC.

We evaluate DECRA on three text classification benchmark datasets under low-resource settings. Extensive experiments show that our model achieves superior performance than advanced baselines, such as LDMAW and CBERT.

The major contributions of this paper are summarized below:
\begin{itemize}
	\item[1)] We first propose a \textbf{D}iversity-\textbf{E}nhanced and \textbf{C}onstraints-\textbf{R}elaxed \textbf{A}ugmentation (\textbf{DECRA}) for Low-Resource Classification (LRC). Experimental results show that our DECRA outperforms the state-of-the-art approach by 3.8\% in the overall score.

	\item[2)] We propose a $k$-$\beta$ augmentation to enhance the diversity in constrained generating. It can improve diversity by expanding the changing scope and enhancing the degree of complexity.

	\item[3)] We propose to use the masked Language Model (LM) loss on original data as a regularization instead of fine-tuning. It helps to relax the constraints, and eventually improve the generalization ability of classifiers.
\end{itemize}
\section{Related Work}
\label{sec:relatedwork}
\subsection{Language model}
Recently, many works have shown that pre-trained Language models (LM) on large corpora can learn common language representations, which is beneficial for downstream Natural Language Processing (NLP) tasks and can avoid training new models from scratch~\cite{mikolov2013efficient,devlin2019bert}. With the development of computing power, the emergence of deep models (i.e. Transformer~\cite{vaswani2017attention}) and the continuous improvement of training skills, the architecture of LM has evolved from shallow to deep. The first generation of LM is designed to learn contextual-free word embedding. They are usually shallow for computational efficiency\cite{mikolov2013efficient,pennington2014glove}. Although these pre-trained embeddings can capture the semantics of words, they have no context and cannot capture high-level concepts in the context. The second generation of LM focuses on learning contextual word embeddings, such as GPT-2~\cite{radford2019language} and BERT~\cite{devlin2019bert}. 
\subsection{Text data augmentation}
Data Augmentation (DA) in text data, different from the image data~\cite{cubuk2019autoaugment,zhong2020random}, is difficult due to the to preserve grammar and semantics. The text augmentation can be divided into the rule-based approaches and the Language Model (LM)-based approaches.

The rule-based approaches mainly augment labeled data with the prior rules~\cite{sato2018interpretable,wei2019eda,guo2020nonlinear}. Some works inject small perturbation into the representation of labeled data~\cite{miyato2016adversarial}. That increases the model's generalization ability. Some works are inspired by the smoothing hypothesis~\cite{szegedy2013intriguing,szegedy2016rethinking}. They propose to use a weight to mix two labeled data into one generated data~\cite{zhang2017mixup,guo2020nonlinear}. Training with both data, it achieves better generalization ability~\cite{berthelot2019remixmatch,archambault2019mixup}. Some works use a pre-trained translation model to augment the labeled data by translating it into another language and translate it back to its original language~\cite{shleifer2019low}.  

The LM-based approaches use the Language Model (LM) to generate diverse data with constraints. The essential operation is to randomly replace some tokens in the labeled data with contextual and label constraints. Easy Data Augmentation (EDA)~\cite{wei2019eda} generates data without label constraints which often results in the label-drift. That will introduce noise into the generated data. Staged fine-tuned LM on labeled data is suitable for the operation~\cite{marivate2020improving}. The Contextual Augmentation (CA)~\cite{kobayashi2018contextual} uses an LSTM based LM to improve the contextual constraints. The LSTM-based LM with context-free word embeddings can not handle the contextual constraints well. Therefore, the contextual aware embeddings are introduced into DA~\cite{wu2019conditional,anaby2020not}. 
\section{Problem formulation}
For a text classification dataset $\{(\textbf{x}_i,y_i)\},i\in[1,N]$, where $\textbf{x}_i\in \mathbb{R}^{T\times V}$ and $y_i\in \mathbb{R}^{C}$, $N$ is the training data size, $T$ is the length of data and $V$ is the vocabulary size, $C$ is the number of classes. In Low-Resource Classification (LRC), the $N$ is very small, such as lower than 40 samples per class. That embodies the needs of Language Model (LM) $g_{\theta_a}$ to generate diverse data to improve the generalization ability of the classifier $f_{\theta_c}$, where $\theta_a$ and $\theta_c$ represent the parameters respectively. The generated data should contain constraints as well as maintain diversity to ensure generalization ability. The formulation of an operation in augmentation is
\begin{equation}
\hat{\textbf{x}}_{i,j} = \phi (\textbf{x}_i;k,\beta,\theta_a),j\in[1,\beta]\,.
\end{equation}
Here, $\phi$ is the operation in augmentation, $\hat{\textbf{x}}_{i,j} \in \mathbb{R}^{T\times V}$ is the $j$-th data generated based on $\textbf{x}_i$, $\beta$ is the number of runs for data generation, $\theta_a$ are the parameters of LM. The generated data $\hat{\textbf{x}}_{i,j}$ has the same label with $\textbf{x}_i$.

The classifier learns the map function of
\begin{equation}
Y = f_{\theta_c}(\textbf{X}),
\end{equation}
where $\{(\textbf{X},Y)\}$ is the joint of $\{(\textbf{x}_i,y_i)\}$ and $\{(\hat{\textbf{x}}_{i,j},y_i)\},j\in[1,\beta]$.
\begin{figure*}[htbp]
	\centering
	\includegraphics[width=0.95\columnwidth]{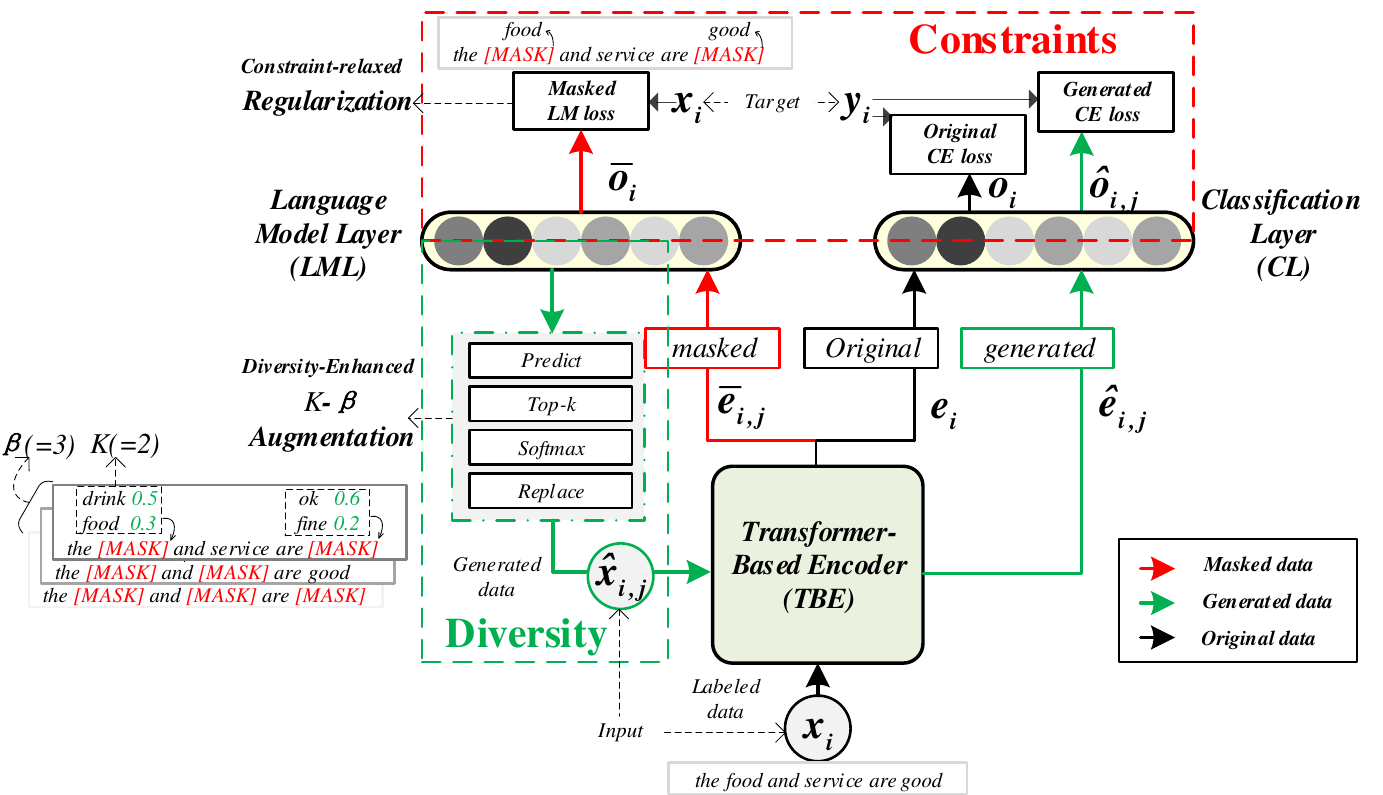}
	\caption{The structure of Diversity-Enhanced and Constraints-Relaxed Augmentation (DECRA).\label{fig2}} 
\end{figure*}
\section{Model Description}
Fig.~\ref{fig2} shows the structure of the \textbf{D}iversity-\textbf{E}nhanced and \textbf{C}onstraints-\textbf{R}elaxed \textbf{A}ugmentation (\textbf{DECRA}). Our DECRA has two essential components based on a backbone model which has a Transformer-Based Encoder (TBE), a Language Model Layer (LML) and a Classification Layer (CL). 
Firstly, the k-$\beta$ augmentation is applied to the original data to generate diversity-enhanced data. 
Secondly, the masked Language Model (LM) loss is introduced as the regularization, which is the relaxed-constraint in generating.

\subsection{Transformer-based encoder}
Transformer-Based Encoder (TBE) stacks multiple layers of transformers~\cite{vaswani2017attention} to encode the text data into embeddings. It is initialized by a pre-trained Language Model (LM) which is trained on large-scale multi-domain datasets. The original data $\textbf{x}_i$ is masked into $\overline{\textbf{x}}_i$. The original data $\textbf{x}_i$ is encoded as follows,
\begin{equation}
\textbf{e}_i = Tansformer_{\theta_t}(\textbf{x}_i)\,.
\label{eq3}
\end{equation}
Here, $\textbf{e}_i \in \mathbb{R}^{T \times H}$ is the embeddings for classification, $Transformer_{\theta_t}$ represents the processing of transformers, $T$ is the length of original data, $H$ is the hidden size of embeddings, $\theta_t$ is the parameters of TBE. Similarly, we can get embeddings $\overline{\textbf{e}}_{i}$ for the masked data $\overline{\textbf{x}}_{i}$.

\subsection{Language model layer}
Language Model Layer (LML) is composed of a fully-connected layer. The fully-connected layer predicts the masked position based on its contextual embedding~\cite{devlin2019bert} that is fundamental for k-$\beta$ augmentation. It also essential to calculate the masked Language Model loss~\cite{devlin2019bert} on original data as a regularization. The $\overline{\textbf{e}}_{i}$ is embeddings of masked data. The prediction is calculated as,
\begin{equation}
\overline{p}_{i} = g_{\theta_{a}}(\overline{\textbf{e}}_{i})\,.
\end{equation}
Here, $\overline{p}_{i} \in \mathbb{R}^{T\times V}$ represents the probabilities of tokens in masked positions, $g_{\theta_{a}}(\cdot)$ maps the embedding size vector to vocabulary size.
\subsection{Classification layer}
Classification Layer (CL) takes the first position of embeddings encoded by the TBE as input, and outputs the class categories. For labeled data, we calculate the predictions as follow,
\begin{equation}
o_i = f_{\theta_c}(\textbf{e}_i)\,,\label{eq5}
\end{equation}
where $f_{\theta_c}(\cdot)$ represents the function of CL, $\theta_c$ is the parameters of CL, $o_i\in \mathbb{R}^{C}$ represents the predictions. 

\subsection{$K$-$\beta$ augmentation} 
$k$-$\beta$ algorithm is designed to enhance the diversity in generating. It aims to augment the original data $\textbf{x}_{i}$ $\beta$ times to get the generated data $\hat{\textbf{x}}_{i,j},j\in[1,\beta]$. 
\begin{equation}
\hat{\textbf{x}}_{i,j} = \phi (\textbf{x}_{i};k,\beta,\theta_t,\theta_a),j\in[1,\beta].
\end{equation}
Here, $\phi$ is the $k$-$\beta$ augmentation, $\beta$ is the set of masks as well as the times of augmentation applied, $k$ is the number of tokens used for replacing the masked position, $\theta_t$ and $\theta_a$ are the parameters in TBE and LML respectively. By this, the changing scope of generated data is expanded. Also, the degree of complexity of generated data is enhanced. 

\begin{figure*}[htbp]
	\centering
	\includegraphics[width=0.6\columnwidth]{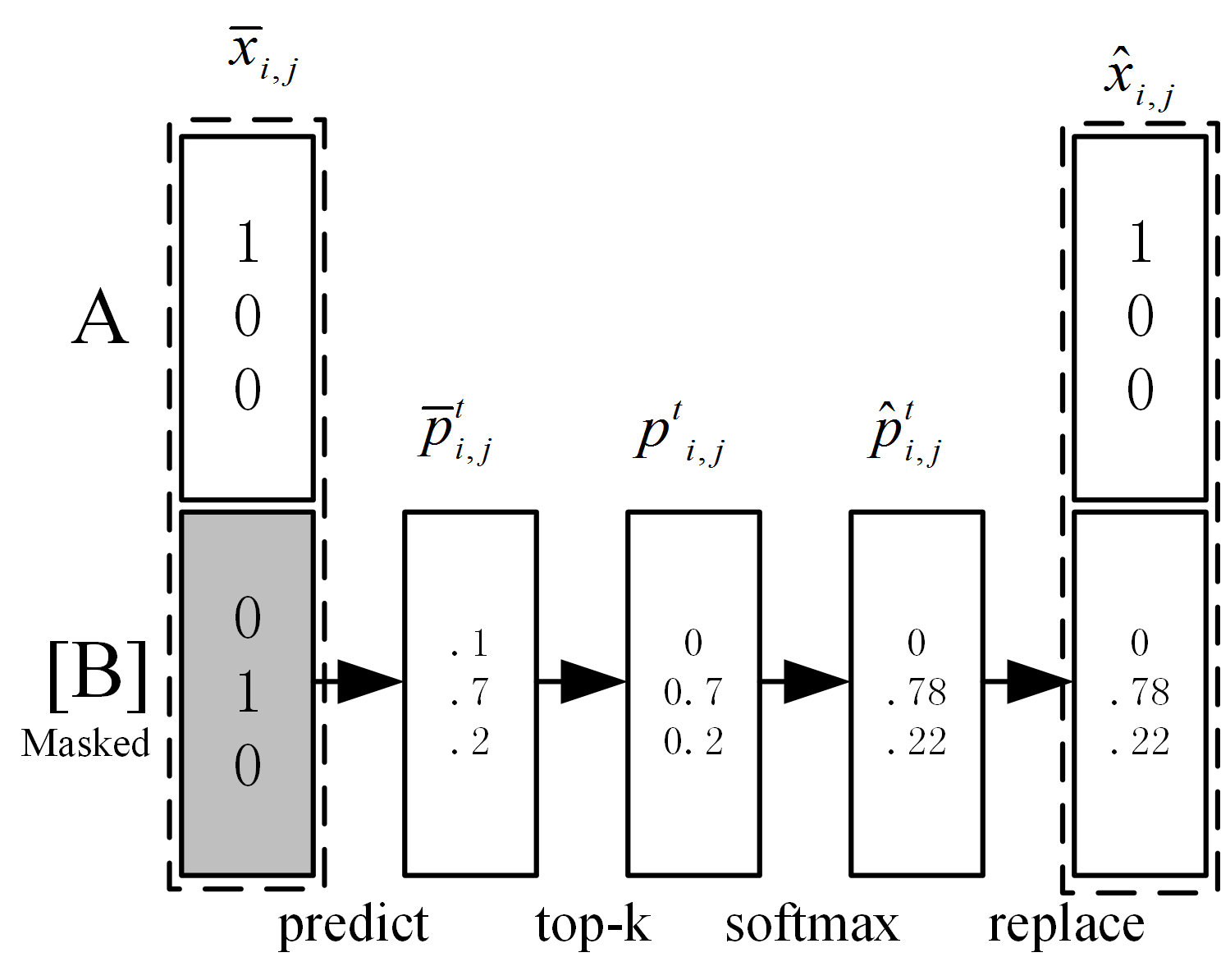} %
	\caption{The demonstration of $k$-$\beta$ augmentation. $V=3$, $T=2$ and $k=2$. For a given sample $\overline{\textbf{x}}_{i,j}$, the masked position is $t=2$, the masked token is \textbf{B}. The $\overline{p}_{i,j}^t$ is generated based on $\overline{\textbf{e}}_{i,j}$, the $p_{i,j}^{t}$ is the top-k of  $\overline{p}_{i,j}^t$, the  $\hat{p}_{i,j}^t$ is the normalized  $p_{i,j}^t$. }
	\label{fig3}
\end{figure*}

For each time of augmentation, the original data $\bf{x}_i$ is randomly masked $\overline{\bf{x}}_i$ for augmentation, as shown in Fig.~\ref{fig3}. Data augmentation consists of four steps: predict, top-k, softmax and replace.

\textbf{\textit{Predict}}. The embedding of the randomly masked position is feed into LML to get the predictions $\overline{p}_{i,j}^{t} \in \mathbb{R}^{V}$. The predictions represent the probabilities of tokens to fit the $t$-th position.

\textbf{\textit{Top-k}}. The top-k sampling, which is often used to improve the diversity in data augmentation~\cite{fadaee2017data}, is used. The top-k probabilities tokens in  $\overline{p}_{i,j}^{t}$ are selected as $p_{i,j}^{t} \in \mathbb{R}^{k}$. 

\textbf{\textit{Softmax}}. The top-k probabilities are feed into a softmax function to normalize the probabilities.
\begin{equation}
\hat{p}_{i,j}^{t} = softmax(p_{i,j}^{t})
\end{equation}
Here, $\hat{p}_{i,j}^{t} \in \mathbb{R}^{k}$ is the normalized top-k probabilities. 

\textbf{\textit{Replace}}. For the convenient of replacement~\cite{hu2019learning}, we fill the value of $\hat{p}_{i,j}^{t}$ into a zero vector to get $p_{i,j}^{t} \in \mathbb{R}^{V}$. Instead of only one sampled token, we use k tokens to replace the masked token $\overline{\textbf{x}}_{i,j}^{t}$, and get the generated data $\hat{\textbf{x}}_{i,j}$. Note that the number of masked tokens is not fixed. The progress is repeated $\beta$ times to get $\hat{\textbf{x}}_{i,j},j\in[1,\beta]$. The labels $\hat{\bf{x}}_{i,j}$ all set to $y_i$ as the setting in ~\cite{wu2019conditional}. The generated data are encoded for classification $\hat{\textbf{e}}_{i,j},j\in[1,\beta]$ as in Eq.~\ref{eq3}. Then, as in Eq.~\ref{eq5}, we can get the prediction of generated data after $k$-$\beta$ augmentation $\hat{o}_{i,j}, j\in[1,\beta]$.

\subsection{Regularization}
Masked Language Model (LM) loss~\cite{devlin2019bert} generates relaxed contextual constraints compared to fine-tuning. The labeled data is corrupted by randomly replacing some positions into maskers. Then, the model learns to predict the original token with the contextual embedding in the masked position. It takes the embeddings of the masked position $\overline{\textbf{e}}_{i}$ as inputs, takes the original tokens $\textbf{x}_i$ as labels, and calculates the loss as follow, 
\begin{equation}
\mathcal{L}_{LM} =\frac{1}{M}\sum_{t=1}^{T}m_t\textbf{x}_i^t \log(f_{\theta_a}(\overline{\textbf{e}}^t_{i}))\,,
\end{equation}
where $\mathcal{L}_{LM}$ represents the masked LM loss, $m_t=1$ indicates the token on $t$ position is masked, $\textbf{x}_i^t\in \mathbb{R}^{V}$ is the $t$-th token, $\theta_a$ is the parameters of LML, $M$ is the number of masked positions.
\subsection{Training process}
As described in Algorithm~\ref{alg1}, the cross-entropy between the predictions and $y_i$ is calculated as
\begin{equation}
\mathcal{L}_{CE} =- \frac{1}{N} \sum_{i=1}^{N} y_i \log(o_i),
\end{equation}
where $N$ is the total number of original data. 

Similarly, we can get cross-entropy loss $\hat{\mathcal{L}}_{CE}$ for the data generated by $k$-$\beta$ augmentation,
\begin{equation}
\hat{\mathcal{L}}_{CE} =- \frac{1}{N} \frac{1}{\beta}  \sum_{i=1}^{N}  \sum_{j=1}^{\beta}y_i log(\hat{o}_{i,j}).
\end{equation}
Here, we average the loss calculated on $\beta$ generated data which can get a more stable improvement~\cite{berthelot2019remixmatch}. 

The final loss is weighted average as follow,
\begin{equation}
\mathcal{L}_{final}=\mathcal{L}_{CE} + \lambda_{a} \hat{\mathcal{L}}_{CE}+ \lambda_{lm}\mathcal{L}_{LM}\,.
\end{equation}
Here, The $\lambda_a$ and $\lambda_{lm}$ are the weights for each loss term.
\begin{algorithm}
\caption{The training algorithm of DECRA.\label{alg1}} 
\begin{algorithmic}
\REQUIRE  corpus $\{(\textbf{x}_i,y_i)\},i\in{N}$, $\lambda_a$ and $\lambda_{lm}$, $\beta$ and $k$.
\STATE Initialize  $\theta_a$ and $\theta_t$ by a pre-trained BERT
\STATE Initialize $\theta_c$
\FOR{$epoch =1,\cdots,M$} 
	\FOR{$i=1,\cdots,N$}
		
		\STATE Masking $\textbf{x}_{i}$ to get $\overline{\textbf{x}}_i$
		\STATE Get embeddings $\bf{e}_i$, $\overline{\bf{e}}$ through $Transformer_{\theta_t}$
		\STATE Calculating the $\mathcal{L}_{LM}$ based on  $g_{\theta_a}(\overline{\textbf{e}}_{i})$ and $\textbf{x}_i$
		\STATE Geting the generated data $\hat{\textbf{x}}_{i,j},j\in[1,\beta]$ through $\phi(\textbf{x}_{i};k,\beta,\theta_t,\theta_a)$
		\STATE Calculating the  $\hat{\mathcal{L}}_{CE}$ based on $f_{\theta_c}(Transformer_{\theta_t}(\hat{\textbf{x}}_{i,j}))$ and $y_i$
		\STATE Calculating the  $\mathcal{L}_{CE}$ based on $f_{\theta_c}(\textbf{e}_i)$ and $y_i$
		\STATE $\mathcal{L}_{final}=\mathcal{L}_{CE} + \lambda_{a} \hat{\mathcal{L}}_{CE}+ \lambda_{lm}\mathcal{L}_{LM}$
		\STATE Update the gradients of $\theta_t$, $\theta_a$ and $\theta_c$ 
	\ENDFOR
\ENDFOR
\end{algorithmic}
\end{algorithm}
\section{Experiments}
\subsection{Experimental settings}
\subsubsection{Dataset}
To evaluate the text augmentation in low-resource classification, we use the same settings in~\cite{hu2019learning}. We evaluate the UABC model based on three benchmark classification datasets, including TREC, SST-5, and IMDB. TREC is to categorize a question into six question types~\cite{li2002learning}. SST-5 is the Stanford Sentiment Treebank with five categories of very positive, positive, neutral, negative and very negative~\cite{socher2013recursive}. IMDB is for binary movie review sentiment~\cite{maas2011learning}, Table~\ref{tab1} summarizes the statistics of the three datasets. For each dataset, we randomly sample 15 small datasets. Each contains 40 samples per class for training and 5 (except SST-5 is 2) samples per class for validation. The models are evaluated on the validation set at the end of each epoch. The optimal model on the validation set is evaluated on the full-size testing set. The mean accuracy of the 15 small datasets used as the final result to evaluate the model performance on each dataset. The average of the mean accuracy on three datasets is the overall score for each model. 
\begin{table}[ht]
\caption{The statistics of datasets. \textit{c}: Number of target classes. \textit{l}: Average sentence length. \textit{Train}: Train set size. \textit{Val}: Validation set size. \textit{Test}: Test set size.}
\centering
\label{tab1}
\setlength{\tabcolsep}{2.5mm}{
\begin{tabular}{cccccc}\hline
Data                      & \textit{c} & \textit{l}   & \textit{Train} & \textit{Val} & \textit{Test} \\\hline
SST-5 & 5 & 19  & 200   & 10  & 2210 \\
IMDB                      & 2 & 252 & 80    & 10  & 2500 \\
TREC                      & 6 & 10  & 240   & 30  & 500 \\\hline
\end{tabular}}
\end{table}
\subsection{Comparison methods}
We compare our model with six methods that can be utilized for Low-Resource Classification (LRC). BERT (base, uncased) for text classification without augmentation~\cite{devlin2019bert} is the \textbf{\textit{baseline}}. Five augmentation methods are listed as follow:
\begin{itemize}
	\item \textbf{EDA}~\cite{wei2019eda} is a recent data augmentation approach containing a set of four text augmentation techniques, including synonym replacement, random insertion, random swap, and random deletion.
	\item \textbf{BT\footnote{We implement the back translation based on MarianMT in Transformers, https://huggingface.co/transformers.}}~\cite{shleifer2019low} translates the labeled data into another language and then translates it back into the original language.
\item \textbf{Mixup}~\cite{jindal2020leveraging} generates out-of-manifold samples through linearly interpolating data representations and their corresponding labels of random sample pairs. 
	\item \textbf{CBERT}~\cite{wu2019conditional} is the latest model-based augmentation that uses a conditional BERT, which is pre-trained on a training set, for augmentation.
	\item \textbf{LDMAW}~\cite{hu2019learning} is the state-of-the-art augmentation that uses reinforcement learning to train both the augmenter BERT and classifier BERT for LRC. 
\end{itemize}
\subsection{Implementation details}
We use BERT-base~\cite{devlin2019bert} to initialize our transformer-based encoder and language model layer, and randomly initialize the classification layer. We use Adam optimization~\cite{kingma2014adam} with an initial learning rate of $2e-5$. The epoch is set to 20 and the batch size is 8 for all datasets. For each minibatch data, we use $k$-$\beta$ augmentation with $\beta=18$ and $k=2$. The weights of losses are $w_a=1$ and $w_{lm}=1.5$. For each experiment, the model is evaluated on the validation set after every training epoch, and the optimal epoch on the validation set is evaluated on the test set.
\subsection{Classification in low-resource condition}
\begin{table*}[ht]
	\caption{DA extrinsic evaluation in low-resource settings. Results are reported as Mean (STD) accuracy on full test set. Experiments are repeated 15 times. $^\dagger$ refers to the results reported in ~\cite{hu2019learning}\label{tab2}}
	\centering
	\setlength{\tabcolsep}{2mm}{
		\begin{tabular}{ccccc}
			\hline
			\multirow{2}{*}{Methods}  & \multicolumn{3}{c}{Datasets}      & \multirow{2}{*}{AVG} \\ \cline{2-4}
			&     SST5(200)      & IMDB(80)      & TREC(240)      &     \\ \hline
			Baseline$^\dagger$~\cite{devlin2019bert}                                               &    33.3$\pm$\small{6.2} & 63.6$\pm$\small{4.4} & 88.3$\pm$\small{2.9} & 61.7                 \\\hline
			EDA~\cite{wei2019eda}                                  &    36.8$\pm$\small{6.1} & 62.8$\pm$\small{6.0} & 86.6$\pm$\small{4.1} & 62.1                 \\
                BT~\cite{shleifer2019low}                                  &    35.8$\pm$\small{4.3} & 66.4$\pm$\small{4.2} & 86.6$\pm$\small{4.3} & 62.8                 \\
			Mixup~\cite{jindal2020leveraging}                                    &    36.0$\pm$\small{4.0}   & 67.3$\pm$\small{5.1} & 88.3$\pm$\small{3.2} & 63.9                 \\
			CBERT$^\dagger$~\cite{wu2019conditional}                                     &    34.8$\pm$\small{6.9} & 63.7$\pm$\small{4.8} & 88.3$\pm$\small{1.1} & 62.3                 \\
			LDMAW $^\dagger$~\cite{hu2019learning}                                &    37.0$\pm$\small{3.0} & 65.6$\pm$\small{3.7} & 89.2$\pm$\small{2.1} & 63.9                 \\\hline
			$\textbf{DECRA}$ (our work)                       &  \textbf{40.3$\pm$\small{3.4}} & \textbf{69.0$\pm$\small{4.0}} & \textbf{89.5$\pm$\small{1.6}} & \textbf{66.3} 
\\\hline
	\end{tabular}}
\end{table*}
Table~\ref{tab2} exhibits the results of all models on three datasets. Our DECRA outperforms all baselines on all three datasets. Firstly, our DECRA can improve the classification performance in LRC from 63.9\% to 66.3\%. When compared with LDMAW and Mixup, our model achieves a higher overall score. That benefits from the effects of our $k$-$\beta$ augmentation which effectively enhances the diversity of generated data. Secondly, our DECRA achieves the highest mean accuracy score on every dataset. The stable improvement may benefit from the expanded changing scope in $k$-$\beta$ augmentation. Thirdly, our DECRA has a smaller parameters-scale than LM-based approaches. When compared with CBERT and LDMAW, our model unifies the augmenter and classifier by reducing nearly half of the parameters. Noticeable that the LDMAW uses reinforcement learning to tune the augmenter(BERT) for the classifier(BERT). Our DECRA improves the overall score by a significant margin. It's 3.8\%  improvements against the LDMAW and 6.3\% improvements against CBERT. The improvement benefits from the improvement of generalization ability. 
\subsection{Ablation study}
\begin{table*}[ht]
	\caption{The ablation results(\%) of DECRA model. Results are reported as Mean (STD) accuracy on full test set. Experiments are repeated 15 times.\label{tab3}}
	\centering
	
	\setlength{\tabcolsep}{2mm}{
		\begin{tabular}{ccccccc}\hline
                \multicolumn{1}{c}{\multirow{2}{*}{ID}}  & \multicolumn{1}{c}{\multirow{2}{*}{$k$-$\beta$}} & \multicolumn{1}{c}{\multirow{2}{*}{Reg.}}&  \multicolumn{3}{c}{Dataset}& \multicolumn{1}{c}{\multirow{2}{*}{AVG}} \\ \cline{4-6}
               \multicolumn{1}{c}{}  &\multicolumn{1}{c}{}  &\multicolumn{1}{c}{} & SST5(200)  & IMDB(80)     & TREC(240)       & \multicolumn{1}{c}{} \\\hline
			1&$\times$    & $\times$   & 33.3$\pm$\small{6.2} & 63.6$\pm$\small{4.4} & 88.3$\pm$\small{2.9} & 61.7 \\
	
			2&$\times$& $\checkmark$ & 33.8$\pm$\small{2.9} & 64.6$\pm$\small{4.4} & 86.5$\pm$\small{3.4} & 61.6 \\
			3&$\checkmark$   &  $\times$  & 36.5$\pm$\small{3.2} & 65.6$\pm$\small{5.0} & 89.0$\pm$\small{1.8} & 63.7 \\
			4&$\times$& $\triangle$ & 38.2$\pm$\small{3.3} & 68.8$\pm$\small{4.1} & 88.4$\pm$\small{3.0} & 65.1 \\
			5&$\checkmark$   & $\triangle$ & 39.0$\pm$\small{5.1} & 68.7$\pm$\small{5.4} & 88.7$\pm$\small{1.9} & 65.5 \\
			6&$\checkmark$   & $\checkmark$ & 40.3$\pm$\small{3.4} & 69.0$\pm$\small{4.0} & 89.5$\pm$\small{1.6} & 66.3 \\\hline
	\end{tabular}}
	\\ $\times$ indicates the component is removed form DECRA, $\checkmark$ indicates the component is added in DECRA.  $k$-$\beta$ is the $k$-$\beta$ augmentation and Reg. is the masked LM loss. $\triangle$ indicates the DECRA is pre-trained with masked LM loss and then finetuned as~\cite{gururangan2020don}.
\end{table*}
To better understand the working mechanism of the DECRA, we conduct ablation studies on all three datasets, as listed in Table~\ref{tab3}. 1) Without augmentation, the $ID2$, which has relaxed constraints compared to $ID4$, results in lower classification accuracy. The results indicate that strong constraints are more effective in LRC without augmentation. 2) With augmentation, the $ID6$, which has relaxed-constraints compared to $ID5$, promotes the overall score from 65.5 to 66.3. The improvement of the overall score in LRC with augmentation mainly from the relaxed-constraints. 3) Besides, the $ID3$ outperforms $ID1$ due to the diversity-enhanced $k$-$\beta$ augmentation. Also, the $ID5$ has a higher overall score (65.5) than $ID4$ (65.1). The results show the effects of $k$-$\beta$ augmentation in LRC, which enhance the diversity of generated data.
\subsection{Importance of diversity and constraints. }
\begin{figure}[htbp]
\centering
\subfigure[Results of different setting of $\lambda_a$.]{
\includegraphics[width=0.45\columnwidth]{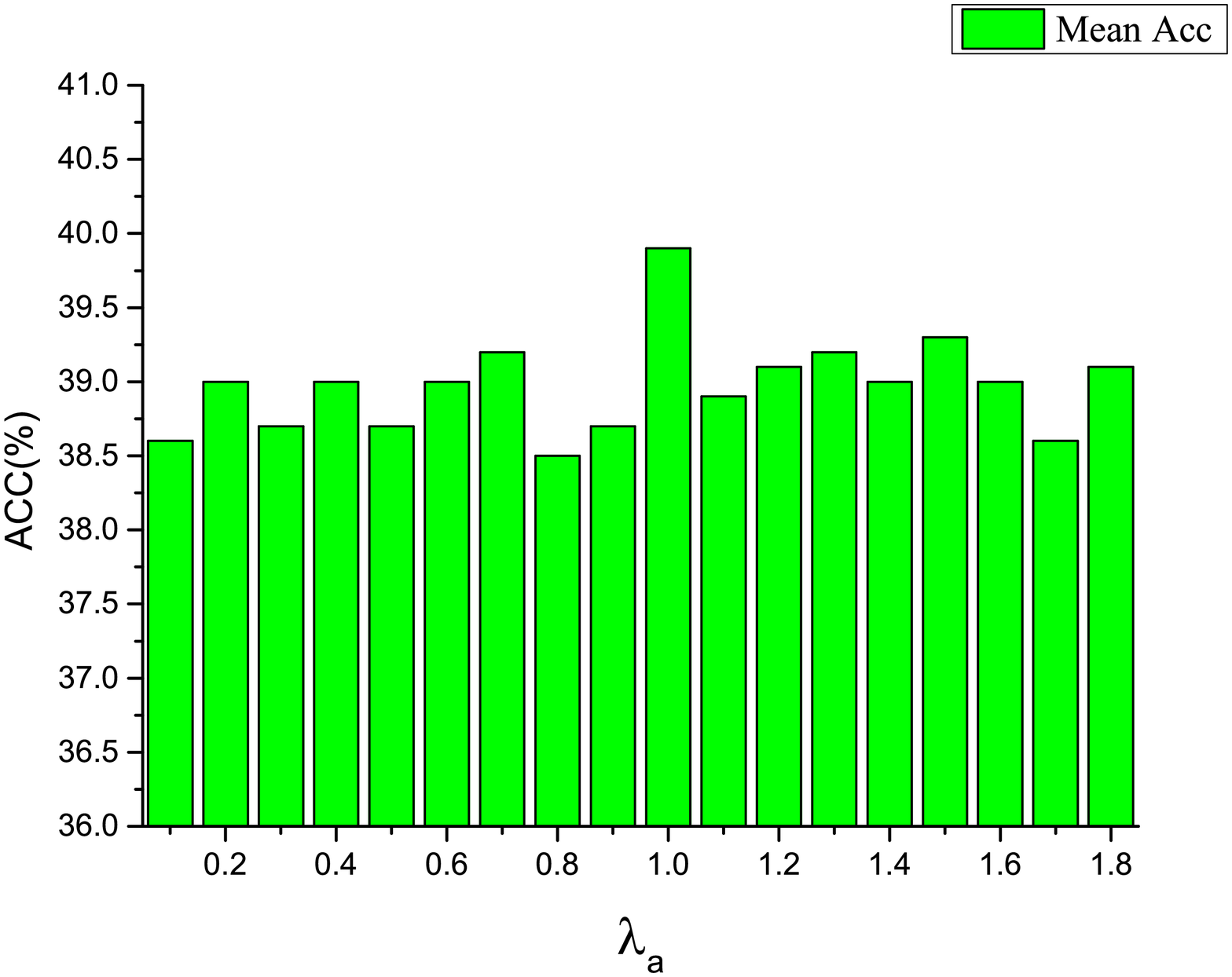}
\label{fig4:subfig1}
}
\subfigure[Results of different setting of $\lambda_{lm}$.]{
\includegraphics[width=0.45\columnwidth]{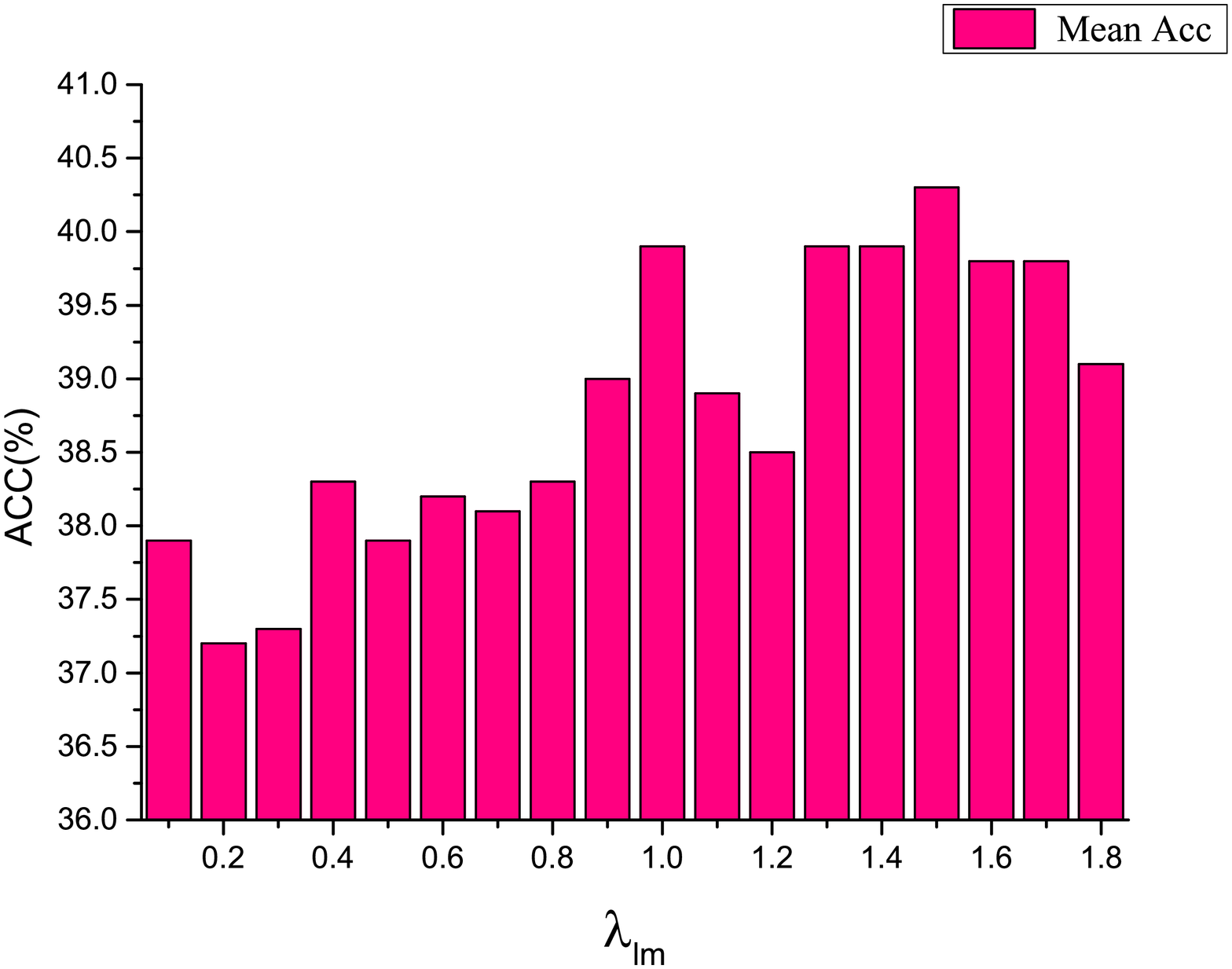}
\label{fig4:subfig2}
}
\label{fig4}
\caption{The results of different loss weights on SST5.}
\end{figure}
\begin{figure}[ht]
	\centering
	\includegraphics[width=0.6\columnwidth]{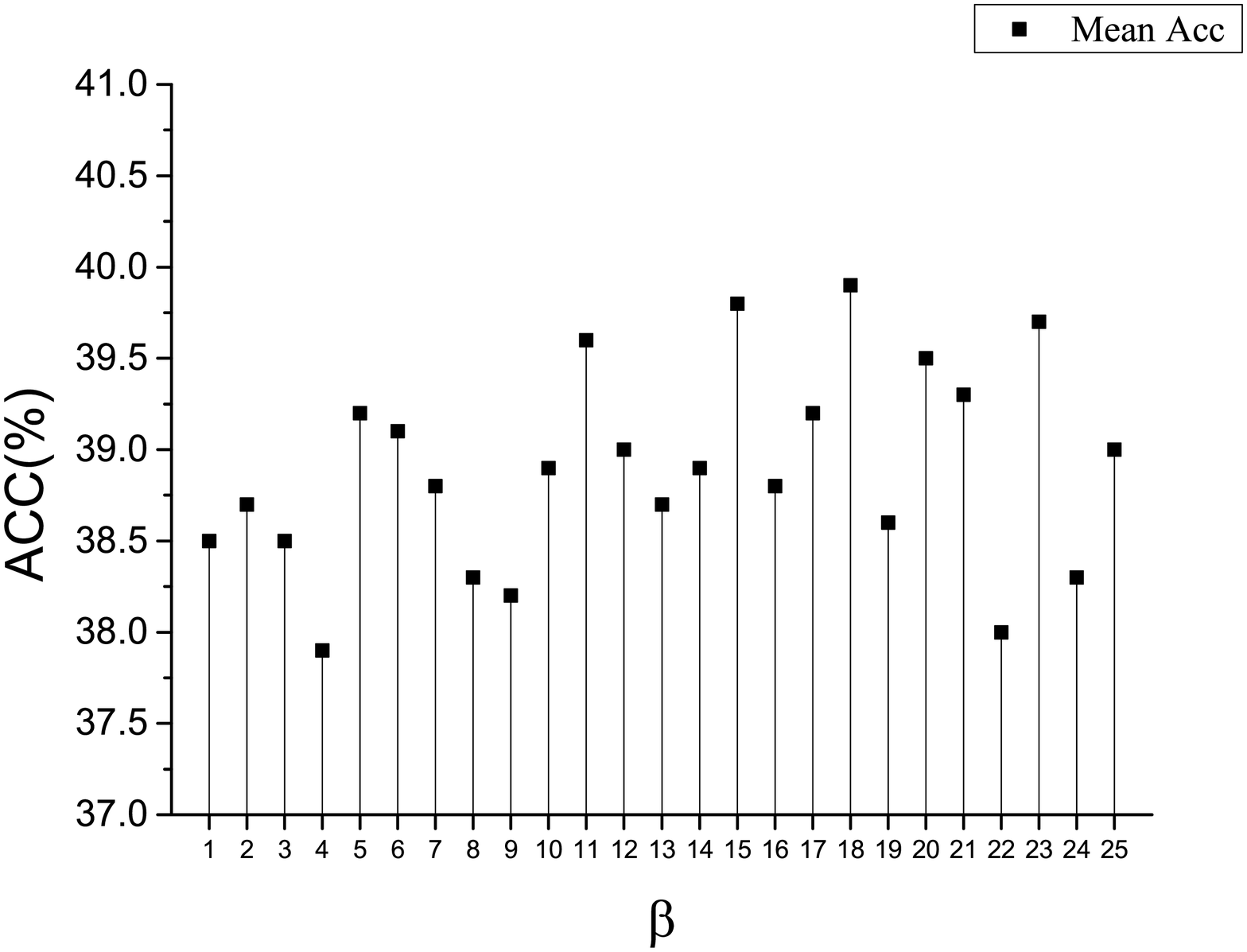} 
	\caption{The results of different $\beta$ on SST-5..}
	\label{fig6}
\end{figure}
\begin{figure}[h]
\centering
\subfigure[]{
\includegraphics[width=0.45\columnwidth]{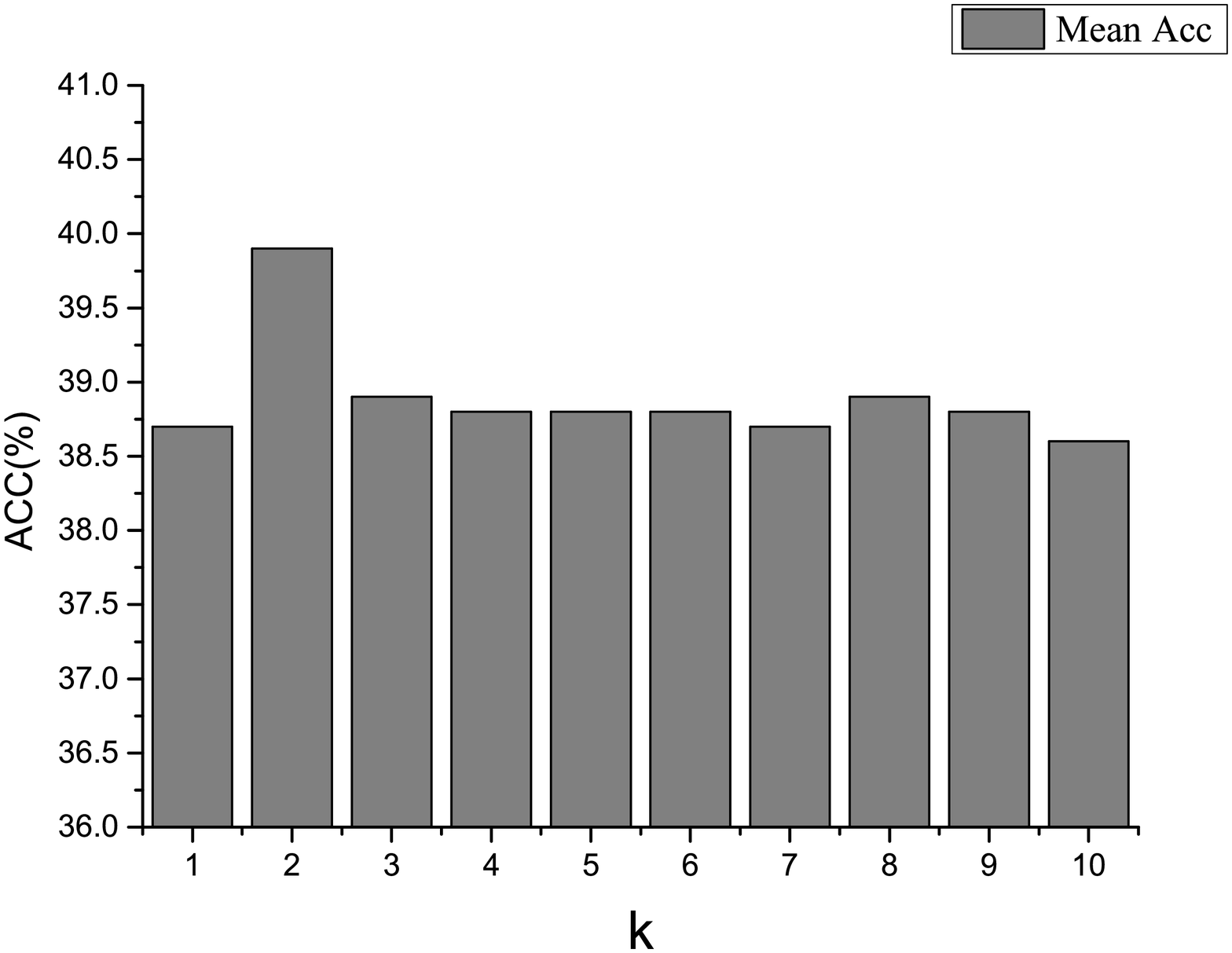}
\label{fig5:subfig1}
}
\subfigure[]{
\includegraphics[width=0.45\columnwidth]{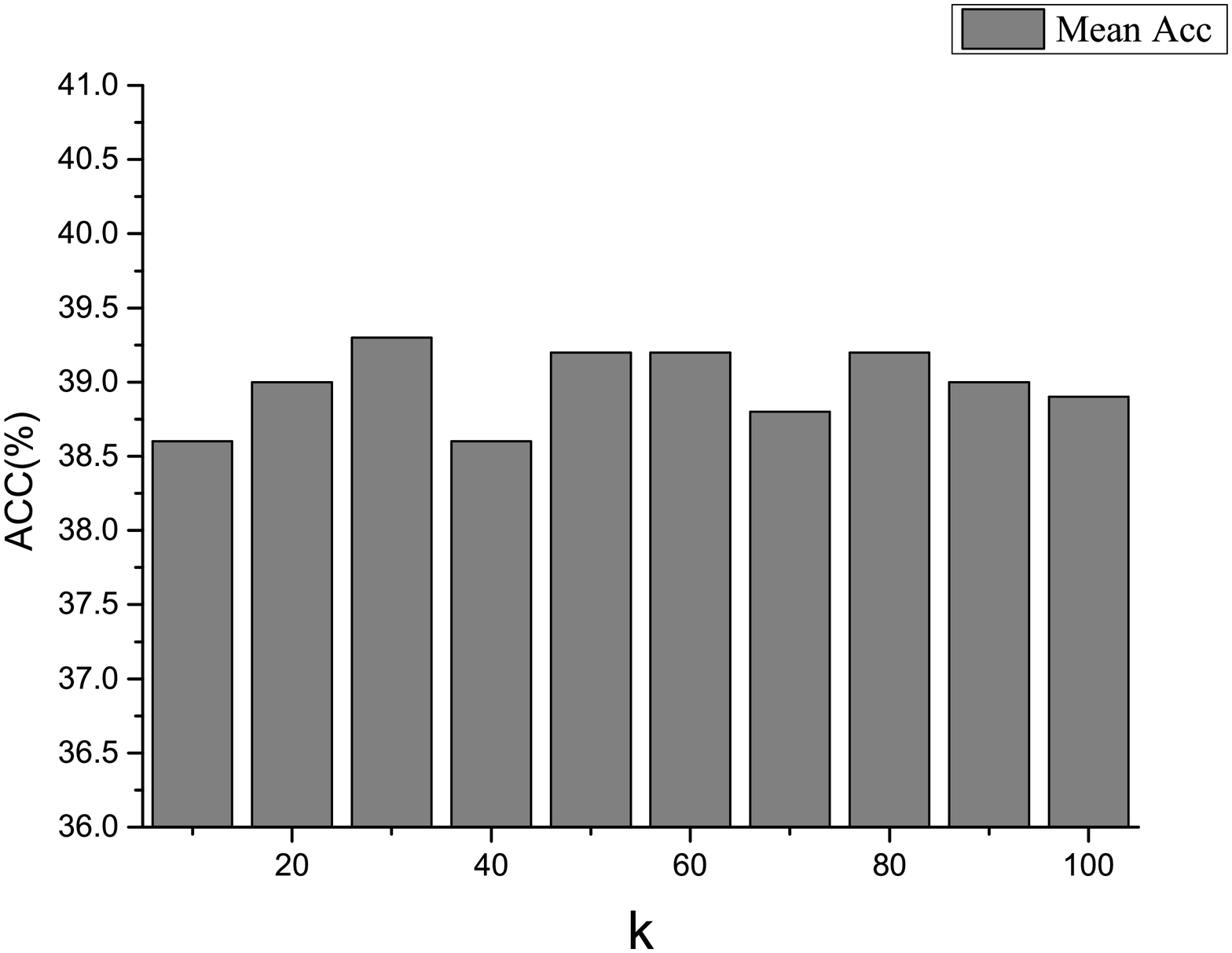}
\label{fig5:subfig2}
}
\caption{The results of different $k$ settings on SST-5.\label{fig5}}
\end{figure}

To analyze the importance of diversity and constraints ($\tilde{\mathcal{L}}_{CE}$ and $\mathcal{L}_{LM}$), we grid search the optimal weights ($\lambda_a$ and $\lambda_{lm}$ ) on SST5. Experiments are repeated 15 times. Firstly, the $\lambda_{mlm}$ is set to $1.0$ in the searching of the $\lambda_{a}$. Then, the $\lambda_{a}$ is set to the optimal ($1.0$) in the search of $\lambda_{lm}$. Fig.~\ref{fig4:subfig1} describes the effects of weight $\lambda_a$ for $k$-$\beta$ augmentation $\hat{\mathcal{L}}_{CE}$. The average accuracy achieves the peak when the $\lambda_a$ is $1.0$. The generated data has equal importance to the original data. This setting is identical to~\cite{hu2019learning}. Fig.~\ref{fig4:subfig2} shows the effects of weight $\lambda_a$ for masked LM loss $\mathcal{L}_{LM}$. The model reaches the optimal classification performance when the $\lambda_{lm}$ is $1.5$. The $\mathcal{L}_{LM}$ has larger weights than $\mathcal{L}_{CE}$. It shows the importance of contextual constraints in LRC.
\subsection{$k$-$\beta$ augmentation on enhancing diversity}
To analyze the effects of the hyperparameter in $k$-$\beta$ augmentation, we conduct two groups of experiments. The $\lambda_a$ and $\lambda_{lm}$ are set to 1.0 as default. Experiments are repeated 15 times.
\subsubsection{Degree of complexity}
To explore the effects of $k$ (degree of complexity) in $k$-$\beta$ augmentation, we conduct two sets of experiments. As shown in Fig.~\ref{fig5}, one set is $k\in[1,2,\cdots,10]$ and another is $k\in[10,20,\cdots,100]$. $k$ is the number of tokens to replace the masked token. In LRC, the operation increases the diversity in generating. Through the top-k operation, the generated data will contain complex information from $(number\ of\ masks)^k$ samples generated by the previous operation in~\cite{wu2019conditional}. As shown in Fig.~\ref{fig5:subfig1}, the model achieves the highest overall score when $k=2$. The mean accuracy decreases along with an increase of $k$. That may be caused by noise information. The model achieves a lower mean accuracy when $k=1$. That is expected as a result of lacking diversity in generating. The results indicate that $k=2$ is an optimal choice for DECRA. 
\subsubsection{Changing scope}
To explore the effect of $\beta$ (changing scope) in $k$-$\beta$ augmentation, we evaluate the model under the setting of $\beta\in [1,25]$ and $k=2$. The $\beta$ is the times of generating maskers for original data. As shown in Fig.~\ref{fig6}, as $\beta$ goes from $1$ to $18$, the mean accuracy of the model tends to increase generally. This benefits from enhancing the changing scope of the augmentation. When $\beta=18$, the highest mean accuracy is achieved for generated diverse data for classification. As the increase of $\beta$ from $18$ to $25$, the model performance begins to fluctuate around $39\%$. The $\beta$ reaches its limit in improving diversity. Therefore, we choose $18$ as the optimal setting.
 
\begin{figure}[h]
\centering
\subfigure[$epoch=10$ and $acc=75.0\%$.]{
\includegraphics[width=0.45\columnwidth]{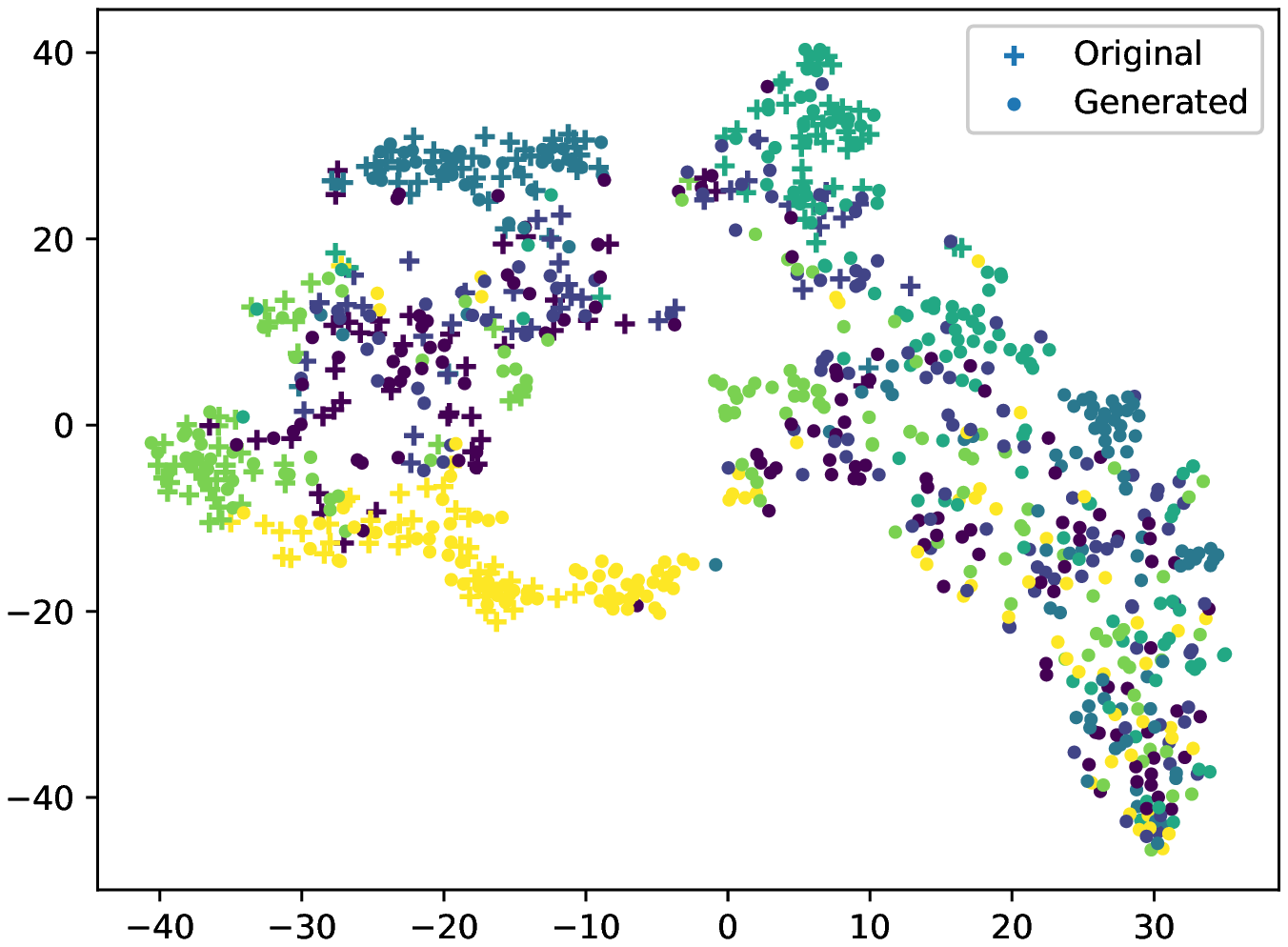}
\label{fig7:subfig2}
}
\subfigure[$epoch=20$ and $acc=88.8\%$.]{
\includegraphics[width=0.45\columnwidth]{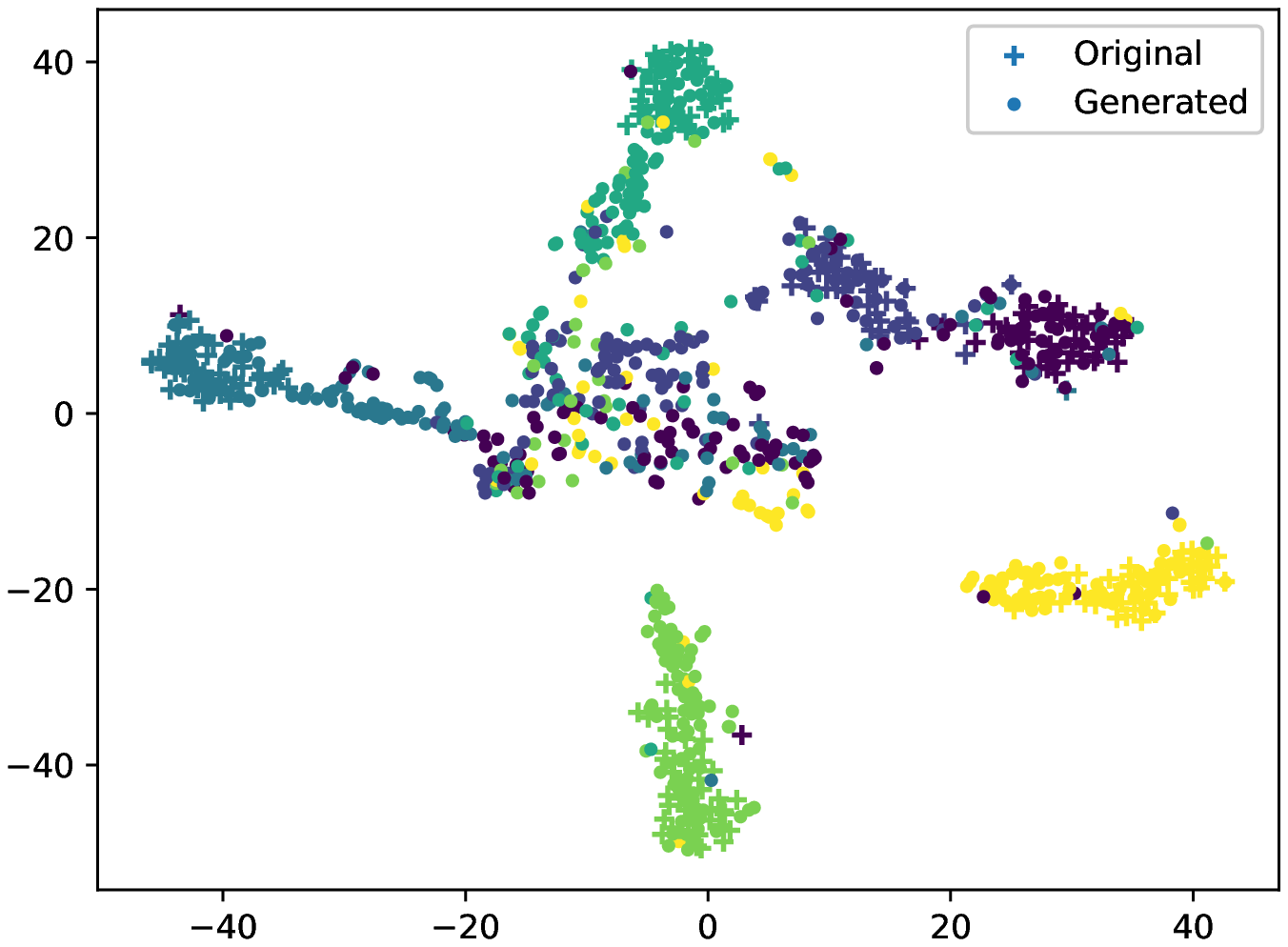}
\label{fig7:subfig4}
}
\caption{The visualization of original data and generated data on TREC ($subset=0$). Different colors are used to mark the original and generated data in each category.\label{fig7}}
\end{figure}
\subsection{Visualization}
To present the effectiveness of the diversity and constraints in DECRA, we visualized the labeled data and generated data in the subset $0$ of TREC with the settings, $k=2$, $\beta=3$, $\lambda_a=1.0$ and $\lambda_{lm}=1.0$. The visualization data in two specific epochs, $epoch =10$ and $epoch=20$, with TSNE~\cite{maaten2008visualizing} represents two different phrases in the training process. As shown in Fig.~\ref{fig7}, we can observe the diversity of the generated data in all training phrases as well as the constraints. The generated data partly distributes around the cluster of its “should be" class and partly distributes distantly away. That demonstrates the effects of diversity and constraints in DECRA. It proves our DECRA works well in LRC.
\section{Conclusion}
In Low-Resource Classification (LRC), the currently used augmentation approaches, such as LDMAW, suffer from generalizing poorly due to strong constraints but weak diversity. To address this dilemma, we propose a Diversity-Enhanced and Constraints-Relaxed Augmentation (DECRA). The DECRA has two essential components on top of a transformer-based backbone model. We propose a $k$-$\beta$ augmentation to enhance the diversity of generated data by expanding the changing scope and enhancing the degree of complexity in generated data. We introduce the masked Language Model loss instead of staged fine-tuning to generate relaxed-constraints. The improved diversity and relaxed constraints help to generate data scattered near or approach the category boundaries. Trained with both labeled and generated data in low-resource conditions, the model achieves better generalization ability. Experimental results demonstrate that our DECRA significantly outperforms state-of-the-art augmentation techniques in low-resource classification. The results may shed some light on LRC.

%
%
%
\bibliographystyle{splncs04}
\bibliography{my}

\end{document}